\documentclass[letterpaper]{article} 
\usepackage[submission]{aaai2026}  
\usepackage{times}  
\usepackage{helvet}  
\usepackage{courier}  
\usepackage[hyphens]{url}  
\usepackage{graphicx} 
\usepackage{natbib}  
\usepackage{caption} 
\usepackage{algorithm}
\usepackage{algorithmic}
\usepackage{newfloat}
\usepackage{listings}
\DeclareCaptionStyle{ruled}{labelfont=normalfont,labelsep=colon,strut=off} 
\floatstyle{ruled}
\newfloat{listing}{tb}{lst}{}
\floatname{listing}{Listing}
\usepackage[utf8]{inputenc} 
\usepackage[T1]{fontenc}    
\usepackage{url}            
\usepackage{booktabs}       
\usepackage{amsfonts}       
\usepackage{nicefrac}       
\usepackage{microtype}      
\usepackage{xcolor}         
\usepackage[most]{tcolorbox}
\usepackage{tikz}
\usepackage{float}
\usepackage{amssymb}
\usepackage{booktabs}
\usepackage{import}
\usepackage{arydshln}
\usepackage{array}
\usepackage{footmisc}
\usepackage{multirow}
\usepackage{makecell}
\usepackage{times}    
\usepackage{xspace}
\usepackage{subfig}
\usepackage[dvipsnames]{xcolor}
\usepackage{amsmath}
\usepackage{amssymb}
\usepackage{booktabs}
\usepackage{caption}
\usepackage{dcolumn} 
\usepackage{graphicx}
\usepackage{tabularray}
\usepackage{multicol}
\UseTblrLibrary{booktabs}
\newtcolorbox[list inside=prompt,auto counter,number within=section]{prompt}[1][]{
    colbacktitle=black!60,
    fonttitle=\small,
    coltitle=white,
    fontupper=\footnotesize,
    boxsep=4pt,
    left=0pt,
    right=0pt,
    top=0pt,
    bottom=0pt,
    boxrule=1pt,
    #1,
}

\newcommand{\et}{\textit{et al.}}
\newcommand{\eg}{\textit{e.g.}}

\newcommand{\Sref}[1]{Sec.~\ref{#1}}

\newcommand{\Fref}[1]{Fig.~\ref{#1}}
\newcommand{\Tref}[1]{Table~\ref{#1}}
\newcommand\mypara[1]{\vspace{1mm}\noindent\textbf{#1}.}

\title{All-in-One Conditioning for Text-to-Image Synthesis}
\author{
     Hirunima Jayasekara\textsuperscript{\rm 1},
    Chuong Huynh \textsuperscript{\rm 1}
Yixuan Ren \textsuperscript{\rm 1}\\
Christabel Acquaye\textsuperscript{\rm 1}
Abhinav Shrivastava\textsuperscript{\rm 1}\\
\textsuperscript{\rm 1}University of Maryland, College Park
}
\affiliations{
    \textsuperscript{\rm 1}University of Maryland, College Park\\

}

\usepackage{bibentry}

\begin{document}

\maketitle

\begin{abstract}
Accurate interpretation and visual representation of complex prompts involving multiple objects, attributes, and spatial relationships is a critical challenge in text-to-image synthesis. Despite recent advancements in generating photorealistic outputs, current models often struggle with maintaining semantic fidelity and structural coherence when processing intricate textual inputs. We propose a novel approach that grounds text-to-image synthesis within the framework of scene graph structures, aiming to enhance the compositional abilities of existing models. Eventhough, prior approaches have attempted to address this by using pre-defined layout maps derived from prompts, such rigid constraints often limit compositional flexibility and diversity. In contrast, we introduce a zero-shot, scene graph-based conditioning mechanism that generates soft visual guidance during inference. At the core of our method is the Attribute-Size-Quantity-Location (ASQL) Conditioner, which produces visual conditions via a lightweight language model and guides diffusion-based generation through inference-time optimization. This enables the model to maintain text-image alignment while supporting lightweight, coherent, and diverse image synthesis.

\end{abstract}

\section{Introduction}
\label{sec:intro}
Recent advancements in text-to-image synthesis have demonstrated improved accuracy and diversity in generating high quality outputs \cite{saharia2022photorealistic,rombach2022high,ramesh2022hierarchical,balaji2022ediffi}. However, current models exhibit significant limitations when processing complex prompts that incorporate multiple objects and attributes with complex spatial relationships. Which leads to synthesized images manifesting inconsistencies with the provided text prompts, including the conflation or complete omission of specified objects, attributes, and spatial arrangements \cite{tong2024mass,bakr2023hrs,chen2024training,phung2024grounded}.

This observation expose a critical gap in the semantic understanding and compositional abilities of existing text-to-image models. While these systems excel at generating individual elements with high fidelity, they struggle to accurately interpret and visually represent the holistic scene described in more elaborate textual inputs \cite{tong2024mass}. Although this behavior is partially inherited from large vision-language models \cite{yuksekgonul2022and,tong2024eyes} such as CLIP \cite{radford2021learning}, which are employed to decode textual input, by introducing sophisticated mechanisms for parsing and maintaining the relational integrity of complex prompt elements may facilitate the embedding of the prompt's comprehensive conceptualization into the image generation process.

Prior work has attempted to address this challenge by constraining image generation through the use of pre-drawn layout maps derived from input prompts \cite{phung2024grounded,li2023gligen}. However, this approach presents several limitations. Primarily, it restricts the placement and dimensioning of objects to the confines of the generated layout, potentially compromising the flexibility and naturalness of the resulting compositions. Furthermore, this method often results in a notable lack of diversity among the generated images. The separate generation of layouts significantly constrains the potential for dynamic adjustments to the spatial structure during the diffusion process, thereby limiting the model's ability to adapt and refine the image composition in response to emerging details or contextual nuances. This rigidity in layout generation may inadvertently constrain the model's capacity to produce varied and contextually appropriate visual representations.

Our work mitigate this problem by grounding the text-to-image synthesis with intermediate conditioning. 
Specifically, we impose an explicit representation on the text prompt using scene graphs and leverage the structuredness it provides to perform noise latent restructuring.
The use of scene graphs in image generation pipelines has been previously explored in the literature. However, existing approaches have predominantly employed scene graphs either to modulate text embeddings derived from the input prompt \cite{feng2022training,shen2024sg,feng2023layoutgpt} or to train auxiliary modules that enforce strict visual constraints for layout guidance\cite{farshad2023scenegenie,wang2024scene}.
We leverage scene graphs to generate soft visual conditions in a zero-shot manner, subsequently use them to guide the image generation process.
To achieve this, we propose Attribute-Size-Quantity-Location(ASQL) Conditioner, which performs two core functions, 1) generating visual conditions via a light weight Large Language Models(LLMs) 2) leveraging the conditions to perform inference-time diffusion optimization\cite{chefer2023attend}. 
We ensure that these conditions comprehensively capture all visual aspects of an image, while adhering to the constraints specified in the textual conditions, thereby enabling the model to generate images with coherence and flexibility.

Through extensive experiments we show that by plugin ASQL Conditioner, our method can significantly improve over state-of-the-art models.
Our key contributions are as follows:
\begin{itemize}
\setlength\itemsep{-0.1em}
\item We propose a zero-shot, robust image generation pipeline that incorporates scene graph-based conditioning to enhance spatial comprehension and diversify generation outcomes, complemented by a novel regime of loss functions that model spatial, size, quantity, and attribute relationships through soft region allocations based on fuzzy logic, enabling more natural object interactions.
\item We introduce a lightweight conditioning framework to generate visual guidance constraints through multi-turn prompting using light weight low-parameter LLMs.
\item Experimental results demonstrate that the proposed method achieves state-of-the-art performance on three standard benchmark datasets, highlighting its effectiveness and robustness across diverse evaluation scenarios.

\end{itemize}

\section{Related Work}

\mypara{Image Generation}
Image generation has progressed significantly from VAEs~\cite{Kingma2014} and GANs~\cite{goodfellow2020generative} to the current dominance of diffusion models~\cite{ho2020denoising, rombach2021highresolution, Ramesh2022HierarchicalTI, peebles2023scalable}. Diffusion models generate images by reversing a noise process, offering high fidelity and controllability. Ho \et~\cite{ho2020denoising} introduced the basic framework, while Song \et~\cite{song2021scorebased} extended it with a score-based approach. Rombach \et~\cite{rombach2021highresolution} improved efficiency by operating in latent space rather than pixel space. More recently, PixArt-$\alpha$~\cite{chen2023pixartalpha} demonstrated that a transformer-based diffusion architecture can scale text-to-image generation to high resolutions while maintaining strong semantic alignment.

\mypara{Conditional Diffusion Models}
Conditional diffusion models allow generation to be guided by labels, text, or layouts~\cite{pmlr-v162-nichol22a, Ho2022ClassifierFreeDG, rombach2021highresolution}. Several approaches enhance layout control by fine-tuning on layout-image pairs~\cite{avrahami2023spatext, li2023gligen}, or modifying sampling~\cite{balaji2022ediffi, bansal2023universal, chen2024training}. Mixture-of-Diffusion~\cite{jimenez2023mixture} and MultiDiffusion~\cite{kumari2023multi} denoise local regions and fuse results, while Dense Diffusion~\cite{kim2023dense} alters attention maps directly. Recent works~\cite{phung2024grounded, feng2024ranni} use LLM-generated layouts but require costly inference and impose rigid constraints, leading to less natural outputs.

Others modify cross-attention maps at test time~\cite{chefer2023attend, phung2024grounded, zhang2024object, sueyoshi2024predicated}. Attend-and-Excite~\cite{chefer2023attend} updates latents using attention, while EBAMA~\cite{zhang2024object} and Predicated Diffusion~\cite{sueyoshi2024predicated} introduce object-attribute consistency losses. We propose a hybrid strategy: identify plausible regions per entity and apply soft spatial guidance, avoiding rigid layouts while maintaining alignment and diversity.

Existing approaches ~\cite{xu2024joint,wu2023imagine,feng2024ranni} rely on extensive training procedures to align scene representations with image generation, often requiring task-specific architectures~\cite{feng2024ranni,farshad2023scenegenie} and supervised data. In contrast, our proposed method is training-free, leveraging pre-trained diffusion models without additional fine-tuning or task-specific retraining, which significantly reduces computational cost while maintaining competitive performance.


\mypara{LLMs for Visual Understanding}
LLMs have shown promise in spatial reasoning~\cite{xu2024evaluating, wei2022chain, yamada2023evaluating}. CoT prompting~\cite{wei2022chain} improves performance on spatial tasks, while recent studies~\cite{xu2024evaluating, yamada2023evaluating} show that even small models can reason about basic spatial relations. These findings motivate our use of lightweight LLMs for extracting soft spatial constraints from text.


\mypara{Scene Graph Generation}
Scene graphs represent objects and their relationships in a structured format, aiding tasks like image retrieval~\cite{schuster2015generating, pham2024composing, johnson2015image} and caption evaluation~\cite{yang2023transforming}. While their use in image generation is limited~\cite{johnson2018image, herzig2020learning}, recent work has explored generating scene graphs from text~\cite{choi2022sgram, zhong2021learning, wang2019on}, enabling structured guidance from unstructured input.
\section{Method}
\label{sec:method}
\begin{figure*}[t]
  \centering
  \includegraphics[width=\textwidth]{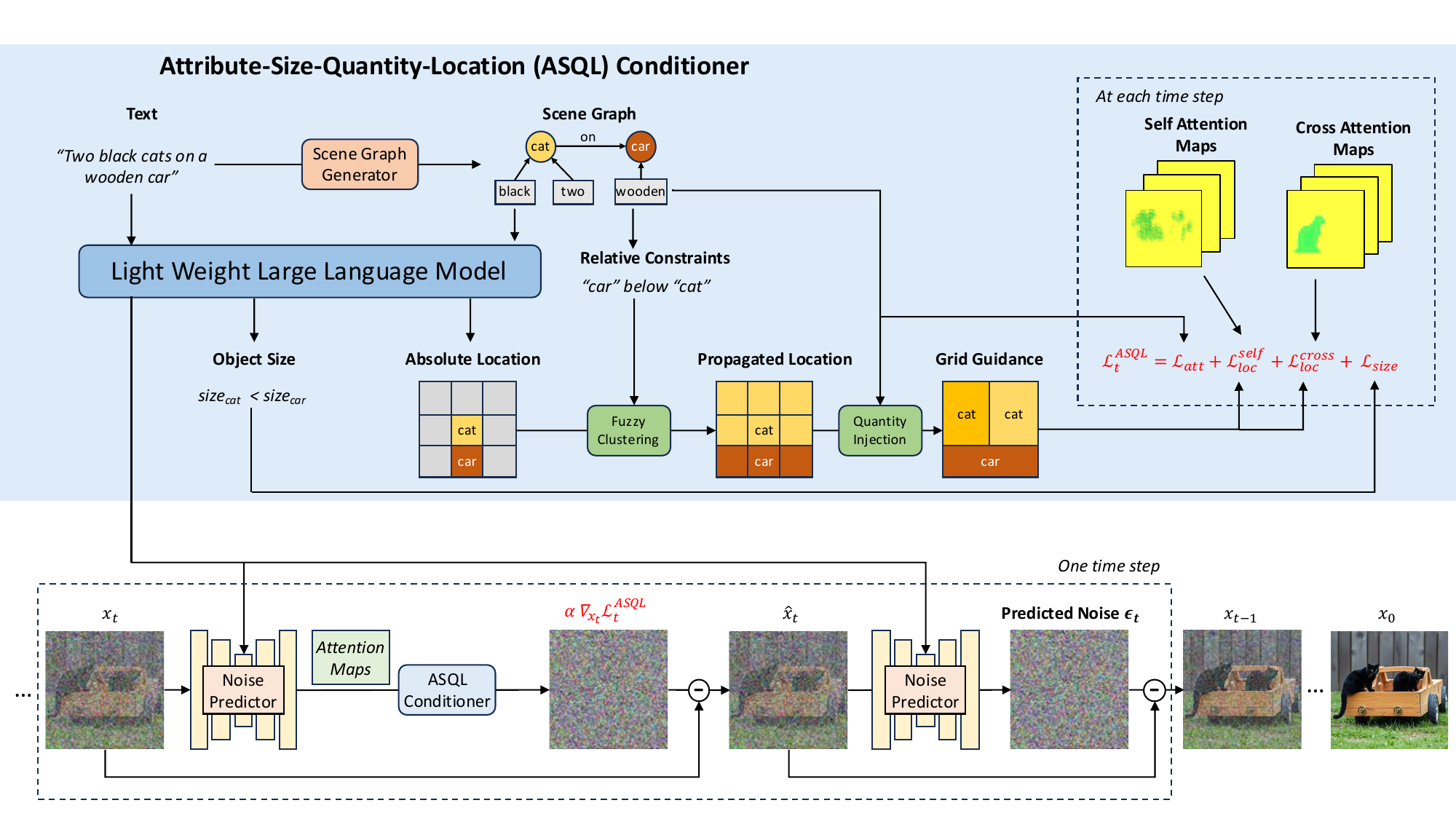}
  \caption{\textbf{Our ASQL Conditioning Pipeline}. Given a text condition, we generate intermediate conditions to control the image denoising process at each time step. All properties (size, quantity, attributes) of each entity and relationships between entities are considered to achive the best result.}
\label{fig:pipeline}
\end{figure*}

In this section, we describe our approach to improve the attribute-size-quantity-location (ASQL) awareness of text-to-image generation pipeline. We first start with the preliminaries of foundations in~\Sref{method:prelim}, followed by the overview of our solution in~\Sref{method:overall}. The details of ASQL conditioning are explained in the last two sections.

\subsection{Preliminaries}
\label{method:prelim}

\mypara{Text-to-image Diffusion Models} Text-to-image diffusion models rely on an iterative denoising process starting from a random Gaussian noise $x_T \sim \mathcal{N}(0, 1)$. At each denoising timestep $t$, a model (\eg, UNet) predicts the noise residual $\epsilon_t = \phi(x_t; \mathbf{c})$ given the noisy input $x_t$ and a conditioning signal $\mathbf{c}$. In text-to-image generation, the condition $\mathbf{c} = f_t(\mathbf{w}) \in \mathbb{R}^{n \times e}$ corresponds to text embeddings obtained from a text encoder $f_t$ (\eg, CLIP) applied to the input text $\mathbf{w} = (w_1, \ldots, w_n)$. Text guidance is incorporated into the denoising process via a cross-attention mechanism, where the \textit{key} $\mathbf{K} \in \mathbb{R}^{n \times d}$ and \textit{value} $\mathbf{V} \in \mathbb{R}^{n \times d}$ are computed from $\mathbf{c}$ using learned linear projections. Given a set of queries $\mathbf{Q} \in \mathbb{R}^{hw \times d}$ derived from the image feature map of size $h \times w$, the cross-attention map $A_t \in [0, 1]^{hw \times n}$ at timestep $t$ is computed as:
\begin{align}
A_t = \text{softmax}\left( \frac{\mathbf{QK}^\top}{\sqrt{d}}\right)
\end{align}
which captures the association between each word token $w_i$ and spatial locations in the image feature map.

\mypara{Inference-time Diffusion Optimization} To enhance text-image alignment, many methods employ inference-time optimization. Given a noisy input $x_t$ at timestep $t$ and an inference-time loss $\mathcal{L}_t$, the model performs a forward and backward pass through the noise predictor $\phi$ to compute the gradient $\nabla_{x_t}\mathcal{L}_t$. The input is then updated as $\hat{x}_t = x_t - \alpha \nabla_{x_t}\mathcal{L}_t$, where $\alpha$ is a controlling strength. The noise predictor then refines its prediction based on the updated input $\hat{x}_t$. The loss $\mathcal{L}_t$ often incorporates attention maps $A_t$ to guide the spatial alignment between the text tokens and image regions. The process is illustrated in the below part of ~\Fref{fig:pipeline}.

\mypara{Scene Graphs} Scene graphs represent images or captions as graphs, where each entity is a node and the relationships between entities are edges. Each node may also contain sub-nodes that represent the entity's attributes. Let $S = (V, E)$ be a scene graph, where $V$ is the set of entities and $E$ is the set of relationships. Each entity $v_i = (o, q, a_1, \ldots, a_k)$ consists of an entity name $o$, the quantity $q$ and a list of corresponding attributes $a_j$. 

\subsection{Overall Framework}
\label{method:overall}







Our proposed pipeline focuses on improving the text-following capacity of text-to-image diffusion models using CLIP text encoder. Following practical approaches~\cite{chefer2023attend,phung2024grounded}, we define a new inference-time loss $\mathcal{L}^{ASQL}$ to modify the input at each time step. Large Language Models (LLM), which has good ability in instruction understanding, help generating guidance in different forms which covers all main challenges about relationship and attribute object binding in text-to-image generation: Attribute, Size, Quantity, Location. 



We leverage LLMs to generate guidance at inference-time. This guidance is then reused at each denoising time step $t$ to compute $\mathcal{L}^{ASQL}_t$ with the attention maps $A_t$. 

\subsection{ASQL Guidance Generation}
\label{method:cond}

The top part of the~\Fref{fig:pipeline} illustrates the pipeline to generate augmented information that adhere the text-to-image generation process. LLMs joins to construct the guidance from the original text and its corresponding scene graph. To further improve the guidance, especially for location and quantity, we propose to enhance the LLM's guidance to have a more accurate grid as a layout constraint.

\mypara{Augmented Guidance Generation} The LLMs takes the caption and its scene graph representation as input and generate two intermediate information pieces to support the conditioning process:

\noindent \textit{Object Size} is a sorted list of objects in size increasing order $\bar{V}=\{v_i | \text{size}(v_i) < \text{size}(v_j)\, \forall i < j\}$. It joins to control the relative size between objects.

\noindent \textit{Absolute Location} is a possible grid location for a given entity in an image grid $G=\{u_{\mathrm{x}\mathrm{y}} \in [0, |V|]\}$ where $u_{\mathrm{x}\mathrm{y}}$ indicates entity id at row $\mathrm{y}-$th and column $\mathrm{x}-$th. Each nonzero value indicates the initial location for the entity.

\noindent \textit{Relative Constraints} contains the relative constraints between objects in both horizontal and vertical directions. While some text explicitly contains this information, others (\eg, ``cat rides a motorbike") does not has. We utilize the relationships derived from each scene graphs to extract these information to improve the visual guidance. For every ordered entity pairs $\{(v_i, v_j)|i\neq j\}$, we compute the vertical and horizontal relation $\mathbf{v}_{ij} \in \{\textsc{above, same, below}\}, \mathbf{h}_{ij} \in \{\textsc{left, same, right}\}$

While the Object Size is directly used to control the image synthesis (explained in~\Sref{method:loss}), the remaining factors contribute to the enhancement process to construct the complete grid guidance. Details regarding prompts and output formats are in the supplementary material.

\mypara{Grid Guidance Enhancement} As illustrated in the top part of~\Fref{fig:pipeline}, the pipeline includes two steps: \textit{(1) Fuzzy Clustering} to assign each cell to an entity and \textit{(2) Quantity Injection} to divide the entity region to $q$ sub-regions. 

In Fuzzy Clustering, we need to assign each cell $(\mathrm{x}, \mathrm{y})$ to the entity with the highest membership:
\begin{align}
    u_{\mathrm{x}\mathrm{y}} = \arg\max_{v_j \in V} \mu \left( v_j, (\mathrm{x}, \mathrm{y})\right)
\end{align}
where $\mu(v_j, (\mathrm{x},\mathrm{y}))$ is the aggregated membership score indicating how suitable it is for entity $v_j$ to occupy cell $(\mathrm{x}, \mathrm{y})$. This membership score is derived by combining the relative spatial constraints between $v_j$ and all other entities $\{v_i| i \neq j\} \in V$, using $\mathbf{v}, \mathbf{h}$ relations. A cell is considered a candidate for $v_j$ only if it satisfies all the directional constraints in the fuzzy rule base. 

In details, for each cell $(\mathrm{x}, \mathrm{y})$, the feasibility of placing $v_j$ with respect to $v_i$ is:

\begin{align}
    \mu_{ij}(\mathrm{x},\mathrm{y})=\mu_{\text{range}_x,ij}(\mathrm{x})\;\cdot\;
             \mu_{\text{range}_y,ij}(\mathrm{y})\in\{0,1\}.
\end{align}
where
\begin{align}
\mu_{\text{range}_x,ij}(\mathrm{x})=
\begin{cases}
1,& \mathbf{h}_{ij}=\textsc{right}\ \land\ \mathrm{x}>\mathrm{x}_i\\
1,& \mathbf{h}_{ij}=\textsc{left}\ \land\ \mathrm{x}<\mathrm{x}_i\\
1,& \mathbf{h}_{ij}=\textsc{same}\ \land\ \mathrm{x}=\mathrm{x}_i\\
0,&\text{otherwise}
\end{cases}
\\
\mu_{\text{range}_y,ij}(\mathrm{y})=
\begin{cases}
1,& \mathbf{v}_{ij}=\textsc{above}\ \land\ \mathrm{y}<\mathrm{y}_i\\
1,& \mathbf{v}_{ij}=\textsc{below}\ \land\ \mathrm{y}>\mathrm{y}_i\\
1,& \mathbf{v}_{ij}=\textsc{same}\ \land\ \mathrm{y}=\mathrm{y}_i\\
0,&\text{otherwise}
\end{cases}
\end{align}

To compute $\mu(v_j, (p,q))$, starting with the all-ones matrix $\mathbf{M}^{(j)}(\mathrm{x},\mathrm{y})=1$, for every other entity $v_i$, we update:
\begin{align}
\mathbf{M}^{(j)}(\mathrm{x},\mathrm{y})\;\leftarrow\;
\mathbf{M}^{(j)}(\mathrm{x},\mathrm{y})\;\cdot\;\mu_{ij}(\mathrm{x},\mathrm{y}).
\end{align}

Because the values are binary, the product is simply the logical \textsc{and} of all pair-wise masks. After looping over all \(i\neq j\) we obtain the final membership
\begin{align}
    \mu\left(v_j,(\mathrm{x},\mathrm{y})\right)=\mathbf{M}^{(j)}(\mathrm{x},\mathrm{y})\in\{0,1\}    
\end{align}

Inorder to incorporate quantity information from the scene graph into the grid $G$ we leverage the quantity annotations provided in the graph. For each entity with $q$ quantities, we divide the respective region into equal $q$ sub-regions, as showed in~\Fref{fig:pipeline}. 

\subsection{ASQL Guidance Loss}
\label{method:loss}




Following previous works~\cite{chefer2023attend,phung2024grounded}, we compute the $\mathcal{L}^{ASQL}$ between the attention score and the guidance obtained from previous section. Instead of using softmax function, we use the score after the sigmoid as:
\begin{align}
    \tilde{A}_t = \sigma \left( \beta \frac{\mathbf{Q} \mathbf{K}^\top}{\sqrt{d}}\right)
\end{align}
where $\beta=100$. The attention still captures the association between each word token and spatial locations but is independent. To simplify, we remove the time step $t$ from the notation of $\tilde{A}$

\mypara{Attribute Guidance Loss} In order to reduce the attribute leakage between entities, we control the attention map between each entity with each corresponding attributes as the attribute loss $\mathcal{L}_\text{att}$:
\begin{align}
    \mathcal{L}_\text{att} = \frac{1}{|V|} \sum_v^V \frac{1}{k} \sum_a^v \text{BCE}\left( \tilde{A}^v, \tilde{A}^a \right) + \eta  \tilde{A}^a ( 1 - \tilde{A}^v)
\end{align}
where $\tilde{A}^o, \tilde{A}^a$ is corresponding attention map of object name and attribute in each entity, BCE is the binary cross-entropy loss, and $\eta$ is the regularization parameter.

\mypara{Size Guidance Loss} To control the size of objects, from the Object Size entity set $\bar{V}$, we compute the pair-wise loss between summation of attention maps of two consecutive object:
\begin{align}
    \mathcal{L}_\text{size} = \frac{1}{|V|}\sum_i^{|V| - 1} \max\left(0, \|\tilde{A}^i\| - \|\tilde{A}^{i+1}\| \right)
\end{align}
where $\tilde{A}^i$ is the corresponding attention map for the $i$-th entity in $\bar{V}$ and $\|.\|$ denotes the summation of all entries in the matrix.

\begin{table*}[t]
\centering
\caption{Performance on all three benchmarks HRS, T2I-CompBench and GenEval. Our achieves the best performance on most metrics. $\dagger,\star, \diamond$ indicates the base as SDv1.4 , PixArt-$\Large{\alpha}$ and SDv2.1 respectively.}
\label{tab:quan}
\scriptsize
\begin{tblr}{width=\linewidth,colsep=2pt,colspec={@{}X[4, l]|*{4}{X[1, c]}|*{4}{X[1, c]}|*{5}{X[1, c]}@{}}}
\toprule
\SetCell[r=3]{c} Method  & \SetCell[c=4]{c} HRS & & & & \SetCell[c=4]{c} T2I-CompBench & & & & \SetCell[c=5]{c} GenEval & & & & & \\ 
\hline
& Count & Spatial & Size & Color & Color & Texture & Shape & Spatial & 2-Obj & Count & Color & Position  & Attribute \\
& F1~$\uparrow$ & Acc~$\uparrow$ & Acc~$\uparrow$ & Acc~$\uparrow$ & BLIP~$\uparrow$ & BLIP~$\uparrow$ & BLIP~$\uparrow$ & Acc~$\uparrow$ & Acc~$\uparrow$ & Acc~$\uparrow$ & Acc~$\uparrow$ & Acc~$\uparrow$ & Acc~$\uparrow$ \\
\hline
CLIP Retrieval\cite{beaumont-2022-clip-retrieval}  & - & - & - & - & - & - & - & -      & 0.22     & 0.37      & 0.62  & 0.03   & 0.00    \\
Mini DALL-E \cite{Dayma_DALL·E_Mini_2021}  & 0.39 & 0.04 & 0.02 & 0.02 & - & - & - & -                & 0.11     & 0.12     & 0.37  & 0.02  & 0.01    \\
\midrule
SDv1.4 \cite{rombach2022high}& 0.58 & 0.09 & 0.09 & 0.13 & 0.37 & 0.42 & 0.37 & 0.13 & 0.38    & 0.35     & 0.76  & 0.04  & 0.06    \\
Attend-n-Excite$^\dagger$~\cite{chefer2023attend} & 0.61 & 0.10 & 0.11 & 0.20  & - & - & - & - &  - & - & - & - & -\\

Ours$^\dagger$ & 0.51 & 0.21 & 0.16 & 0.14 & 0.41 & 0.44 & 0.39 & 0.20   &   0.40     &  0.31    & 0.74   & 0.16   &  0.09    \\
\midrule
PixArt-$\Large{\alpha}$  \cite{chen2023pixartalpha}   & 0.55  & 0.19  & 0.16  & 0.18 & 0.39  & 0.45  &  0.36 & 0.18 & 0.47     & 0.42    & 0.78  & 0.08  & 0.12 \\
Ours $^\star$ & 0.55    & 0.21  & 0.18  & 0.20 & 0.41  & 0.46 & 0.39 &  0.21 & 0.51  & 0.42  & 0.80  & 0.12 & 0.08 \\
\midrule
SDv2.1 \cite{rombach2022high}& 0.56 & 0.17 & 0.13 & 0.22 & 0.57 & 0.50 & 0.45 & 0.17 & 0.51     & \textbf{0.44}     & \textbf{0.85}  & 0.07  & 0.17     \\
Composable$^\diamond$~\cite{liu2022compositional}  & - & - & - & - & 0.41    & 0.37    & 0.33    & 0.08 & - & - & - & - & -\\
Structured$^\diamond$~\cite{feng2022training}  & - & - & - & - & 0.50    & 0.49    & 0.42    & 0.14  & - & - & - & - & -\\
Attend-n-Excite$^\diamond$~\cite{chefer2023attend}  & 0.63 & 0.26 & 0.14 & 0.31 & 0.64 & 0.60 & 0.45 & 0.15  & 0.65 &  0.38 & 0.80 & 0.12 & 0.21   \\
EBAMA$^\dagger$~\cite{zhang2024object} & 0.63 & 0.27 & 0.12 & 0.32  & - & - & - & - & 0.67 & 0.36 & 0.84 & 0.12 & 0.23 \\
Ours$^\diamond$  &  \textbf{0.66} & \textbf{0.28} & \textbf{0.17} & \textbf{0.37} & \textbf{0.69} & \textbf{0.63} & \textbf{0.52} & \textbf{0.27} &   \textbf{0.69}    &   0.40    &  0.76  &\textbf{0.41}&\textbf{0.24}    \\
\bottomrule
\end{tblr}
\end{table*}

\mypara{Location Guidance Loss} The location-based penalization comprises two components. First, we calculate a location loss $\mathcal{L}_\text{loc}^\text{cross}$  derived from cross-attention maps. We employ grid guidance in the form of binary masks $G$. To mitigate the effects of hard masking, we apply a Euclidean distance transform \cite{huang2002euclidean} to each entity mask corresponding to entity $v$, resulting a softened representation denoted as $\bar{G}^o$. To ensure holistic representation of each object, $\mathcal{L}^\text{cross}_\text{loc}$ is computed using the 2-D Dice loss\cite{sudre2017generalised} for each entity $v$:
\begin{align}
    \mathcal{L}_\text{loc}^\text{cross} = \sum_e^V 1 - \frac{2 \|\tilde{A}^v \cdot \bar{G}^v\| }{\|\tilde{A}^v\| + \|\bar{G}^v\|}
\end{align}
Secondly, the self-attention-based location loss $\mathcal{L}_\text{loc}^\text{self}$ is computed. Similarly, we compute a Euclidean distance–transformed 3D mask $\hat{G}^v$ for each entity $v$ using $G$. where, $\bar{G}^v$ is extended along the channel dimension and multiplied by its flattened counterpart. This operation accentuates spatial regions likely to contain the object, facilitating more precise localization. Subsequently, a 3D Dice loss is computed between the self-attention output $\hat{A}$ and the computed mask. Pseudocode for the mask generation process is provided in the supplementary material.
\begin{align}
    \mathcal{L}_\text{loc}^\text{self} = \sum_e^V \sum_s^{H\times W} 1 - \frac{2 \|\hat{A} \cdot \hat{G}^v\| }{\|\hat{A}\| + \|\hat{G}^v\|}
\end{align}

The final $\mathcal{L}^{ASQL}_t$ loss at timestep $t$ is computed as a weighted combination of all three losses.
\begin{align}
\mathcal{L}^{ASQL}_t=\lambda_1 \mathcal{L}_\text{att}+\lambda_2 \mathcal{L}_\text{size}+\lambda_3\mathcal{L}_\text{loc}^\text{cross}+\lambda_4\mathcal{L}_\text{loc}^\text{self}
\end{align}
where $\lambda_1, \lambda_2, \lambda_3, \lambda_4$ are hyper-parameters.



\renewcommand{\arraystretch}{1.3}
\renewcommand{\tabcolsep}{6pt}

\section{Experiments}
\label{sec:experiment}

\subsection{Baselines and Benchmarks}

\mypara{Baselines} We compare our proposed pipeline with other baselines with stable diffusion versions: Attend-and-Excite~\cite{chefer2023attend}, EBAMA~\cite{zhang2024object}.
We refer to several baselines models while evaluating the capacity of the new model. We compare the proposed model with models without region suggestions given, Stable Diffusion \cite{rombach2021highresolution} (1.4V, 2.1V) and Attend-and-excite \cite{chefer2023attend}, EBAMA \cite{zhang2024object} and multi diffusion\cite{kumari2023multi}.

\mypara{Evaluation Metrics} We utilize HRS \cite{bakr2023hrs}, GenEval \cite{ghosh2023geneval} and T2I-CompBench \cite{huang2023t2i} benchmarks on evaluating the proposed model on several categories: accuracy, robustness, and generalization. For HRS, we focus on the accuracy improvements, we consider four categories, counting, spatial, color and size compositions. Each sample consists of prompt, object’s name and corresponding category used for evaluation. T2I-CompBench  deploys a BLIP-VQA\cite{li2023blip} model to evaluate the color, position, and attribute binding for generated images while deploying an object detector to compute spatial accuracy. Similarly, GenEval employs a object detector followed by a color classifier to identify two object relations, counting, color, position(spatial) and attribute binding.


\mypara{Implementation} We conduct all the experiments on one NVIDIA RTX A6000 GPU with batch size of 1. The proposed pipeline is built on the public official implementation of Attend-and-Excite~\cite{chefer2023attend}. For scene graph generator, we utilize our own scene graph-LLM, finetuning Llama-3.1-Tulu-3-8B~\cite{lambert2024tulu3} on FACTUAL~\cite{li-etal-2023-factual} dataset. Performance of our new scene graph generator and finetuning details is in the supplementary material. We utilized Phi3-instruct\cite{abdin2024phi} as our light weight LLM. We followed the SD basline for sampling which requires in $24\time24$ attention resolution for SD v2.1 and $16\time16$ attention resolution for SD v1.4. We sample the noise latent for 50 steps. Similarly, we used PixArt-$\alpha$~\cite{chen2023pixartalpha} sampling scheme with $64\time64$ attention resolution and 20 sampling steps to generate the final image.
Empirical analysis on hyper-parameters for each pipeline can be found in the supplementary.
\begin{figure*}[t]
    \centering
    \includegraphics[width=0.8\linewidth]{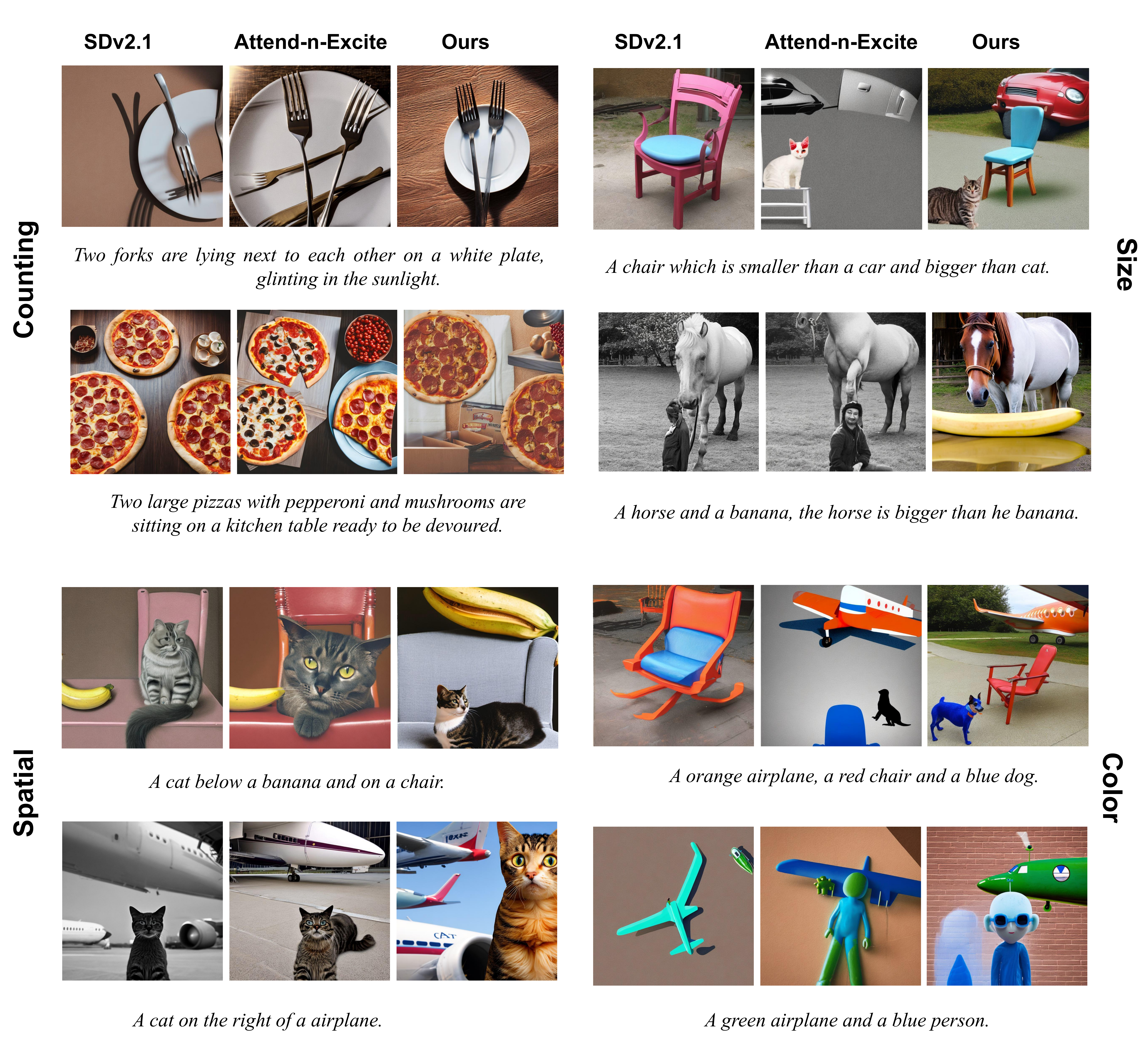}
    \caption{Qualitative results for proposed method. First two columns: images generated from baselines SDv2.1 and Attend-n-Excite with SDv2.1 base, Third column: image generated from proposed method.}\label{tab:quality}
\end{figure*}

\subsection{Quantitative Results}

Comparison on all benchmarks is shown in~\Tref{tab:quan} with both base SD versions. In overall, we achieve the best score in most metrics with a large margin, proving the effectiveness of our solution. In HRS, the proposed method improves the F1 score by 3\% and increases color accuracy by 5\%. Additionally, it achieves a 1\% performance gain in both size and spatial accuracy.

As shown in Table \Tref{tab:quan}, Compared to the baselines, the proposed model yields a 5\% and 7\% increase in BLIP scores for color and shape, respectively, and a substantial 12\% improvement in the spatial category.  

Furthermore, the proposed model demonstrates significantly enhanced performance, achieving a 29\% increase in accuracy for the position category on the GenEval benchmark. Additionally, the model yields accuracy gains of 2\% and 1\% in the two-objects and attributes categories, respectively. However, a decline in performance is observed for the counting and color categories, where as SDv2.1 outperforms latent modification methods.


\subsection{Qualitative Results}

Figure \ref{tab:quality} illustrates qualitative results of the proposed method. We evaluate the model on four sub categories. Namely, counting, spatial, color and size. We present three samples, one each from basline models and proposed model where, Column 1: Stable Diffusion v2.1 , column 2: Attend and Excite with SDv2.1 base, column 3: proposed method. For each example, ground-truth prompt positioned at the bottom.  
Results of our model showcase the ability to bind attributes to respective objects while maintaining spatial consistency across the generated image.
\begin{figure*}[t]
    \centering
    \includegraphics[width=0.8\linewidth]{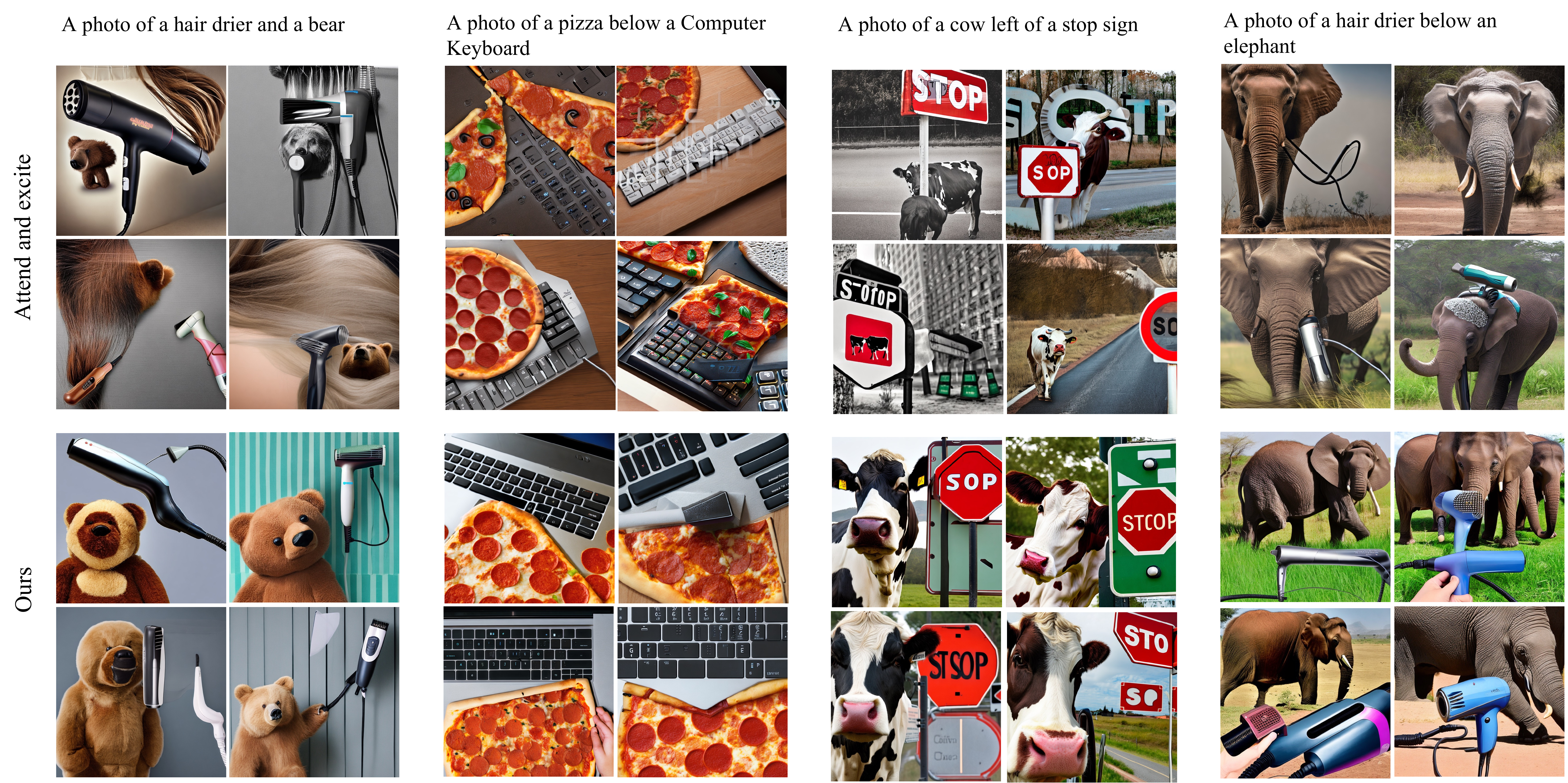}
    \caption{Qualitative results for proposed method. We compare our results on two objects and position with Attend-n-Excite with SDv2.1 base.}\label{tab:div}
\end{figure*}

\begin{figure*}[h]
    \centering
    \includegraphics[width=\linewidth]{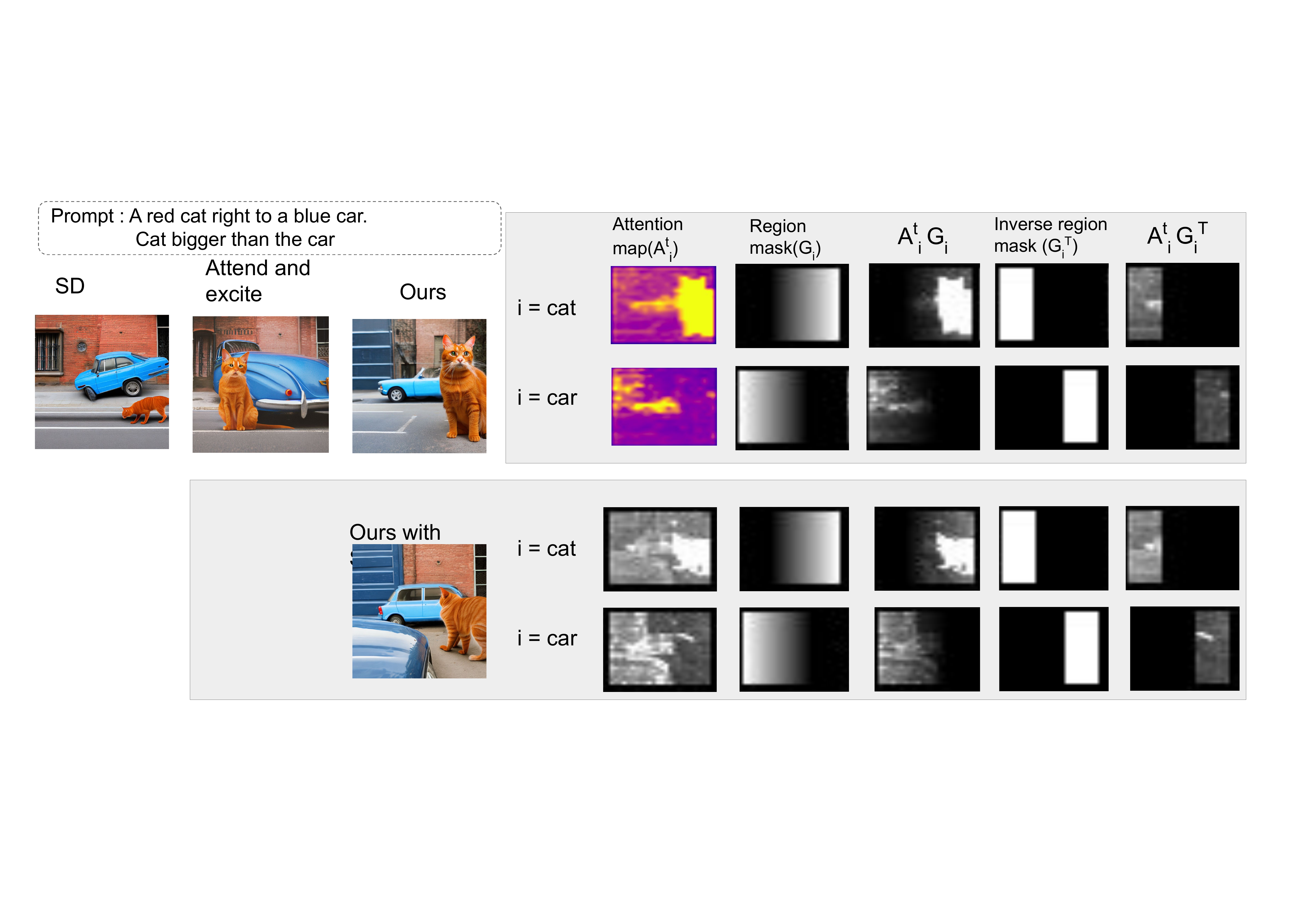}
    \caption{Example of Soft Region masking and effect of dice loss and size loss. First column shows the generated image from SD and Attend and excite. Second column illustrate the image generated with proposed pipeline.}\label{fig:mask}
\end{figure*}
\subsection{Ablation Experiments}
We conduct several ablations on proposed method to illustrate the effectiveness and robustness of proposed modifications.

\begin{table}[t]
\caption{Ablation Experiments on GenEval benchmark. We systematically exclude each loss function and evaluate the model's performance using the remaining components.}
\fontsize{8}{9}\selectfont
\centering
\small{
\resizebox{\linewidth}{!}{%
\begin{tabular}{@{}lcccccc@{}}
\toprule
Effect  & \makecell{Two \\objects}  & \makecell{Counting }   & \makecell{Color }& \makecell{Position} & \makecell{Attribute \\Binding}\\
\midrule
\textit{(0)} Basline & 0.69     &   0.40    &  0.76  &0.41 & 0.24 \\
\textit{(1)} W/O LLM Region Suggestions    &  0.66    &  0.34     &  0.72  &0.34 &0.20 \\

\textit{(2)} Spatial Loss + Size Loss      &  0.66    &   0.39    &  0.74 & 0.40 & 0.20\\
\textit{(3)} Spatial Loss + Attribute Loss     &  0.67    &   0.39    &  0.75 & 0.40& 0.25\\
\textit{(4)} Size Loss + Attribute Loss     &  0.69    &   0.37    &  0.74  &  0.35 & 0.23 \\
\bottomrule
\end{tabular}
\label{tab:aba}
}
}%
\end{table}

\begin{table}[t]
\centering
\small
\caption{Effect of different LLMs in ASQL Conditioner. We utilized smaller variation of Phi, Llama and Deep-seek to compute the visual constrains.}
\resizebox{\linewidth}{!}{%
\begin{tabular}{@{}lccccc@{}}
\toprule
Method                            &\makecell{Two \\objects}  & \makecell{Counting }   & \makecell{Color }& \makecell{Position} &\makecell{Attribute \\Binding}\\
\midrule
Phi3-mini-4k-instruct\cite{abdin2024phi}         & 0.69     &   0.40    &  0.76  &0.41 & 0.25 \\
Llama-3.1-8B-Instruct \cite{touvron2023llama}              & 0.43 & 0.35 & 0.75  &  0.35 &  0.17              \\
\makecell{DeepSeek-R1-Distill-Qwen-1.5B} \cite{guo2025deepseek} &  0.35& 0.25 &  0.52  & 0.20  &0.10\\

\bottomrule
\end{tabular}
}
\label{tab:llm}
\end{table}

\noindent\textbf{Effect of Spatial Loss}: To assess the impact of the spatial loss, we evaluate the model using a combination of size and attribute losses, as presented in Table \ref{tab:aba}\textit{(4)}. The results indicate a significant decline in position accuracy when the spatial loss is omitted.

\noindent\textbf{Effect of Size Loss}: To evaluate the contribution of the size loss, we assess the model's performance using only the spatial and attribute losses. As shown in Table \ref{tab:aba}\textit{(3)}, this results in a decline in counting accuracy. While the size loss primarily ensures that the relative sizes of entities are preserved, it also influences entity presence by encouraging the model to maximize the spatial area allocated to each entity.

\noindent\textbf{Effect of Attribute Loss}: To assess the impact of the attribute loss, we evaluate the model using only the spatial and size losses. As shown in Table \ref{tab:aba}\textit{(2)}, we can observe a drop in attribute accuracy.

\noindent\textbf{Effect of Soft Region Masking}: Figure \ref{fig:mask} illustrate the effect of soft region masking where First column shows the generated image from SD and Attend and excite. Second column illustrate the image generated with proposed pipeline. Top image of the second column is generated with dice loss while the bottom image is generated with size loss. Last Five columns denotes the extracted attention maps for each entity ($A^t_i$), soft region mask for each object ($G_i$), region-entity attention maps ($A^t_iG_i$), inverse region mask ($G_i^T$) and restrictive regions for each entity ($A^t_iG_i^T$) respectively.

\noindent\textbf{Effect of LLM usage}: Proposed method relies on visual information constraints generated by a LLM. Consequently, it is essential to identify a lightweight LLM that balances low computational requirements with high accuracy and coherence in output generation. Through empirical evaluation, we found that Phi-3-mini-4k-instruct \cite{abdin2024phi} is both convenient to deploy and effective in producing consistently structured outputs. To assess the impact of the choice of LLM, we compared the performance of Phi-3 with LLaMA-3.1-8B-Instruct \cite{touvron2023llama} and DeepSeek-R1-Distill-Qwen-1.5B \cite{guo2025deepseek}, as shown in Table \ref{tab:llm}.

\noindent\textbf{Model Capabilities on Diversify Image Generations}:
Figure \ref{tab:div} illustrates qualitative results of the proposed method with Geneval benchmark. We evaluate the model on four sub categories. Namely, counting, spatial, color and size. We generate four samples for each prompt while each generation having different seed. First two rows illustrate the images generated with Attend and Excite with SDv2.1 base and last two rows shows images generated with proposed method. For each example, ground-truth prompt positioned at the top. Results of our model showcase the ability to bind attributes to respective objects while maintaining spatial consistency across the generated image.

\begin{figure}
    \centering
    \includegraphics[width=\linewidth]{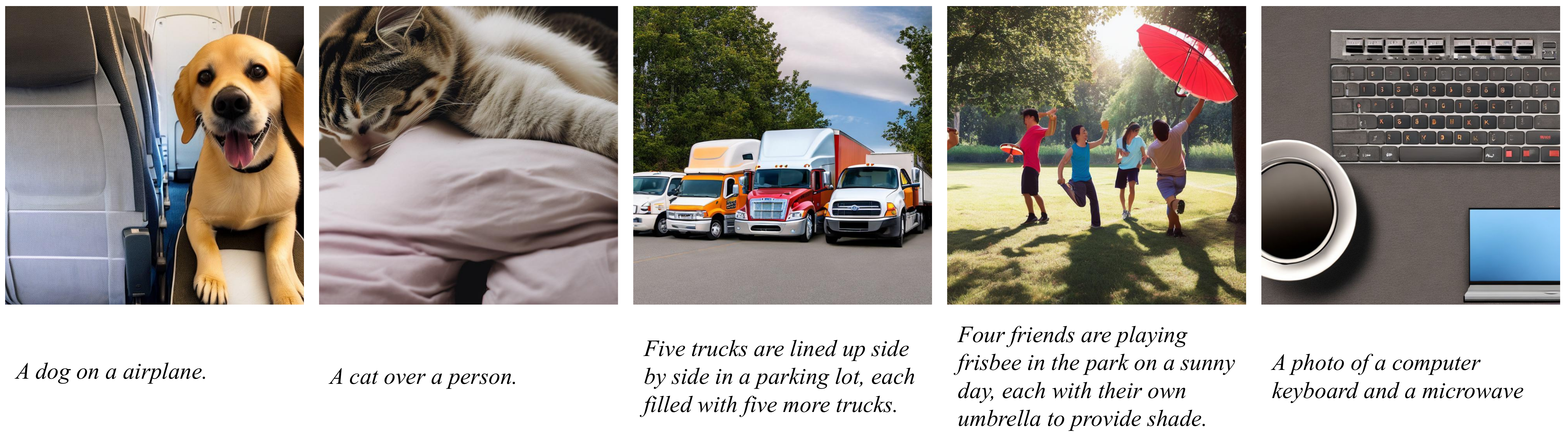}
    \caption{\textbf{Failed Generations.} First two columns: images generated overly literal manner, Last three columns: entity is either absent or too complex to be fully captured by the text prompt.}\label{tab:wrong}
\end{figure}
\subsection{Limitations}
Some failures can be attributed to the model's tendency to interpret prompts in an overly literal manner. As shown in the first two columns of Table \ref{tab:wrong}, these outputs were deemed incorrect, despite accurately reflecting a literal reading of the input descriptions. Despite the proposed model yielding superior performance, it struggles to accurately follow counting instructions when the quantity value increases to higher numbers, thereby resulting in deviations from the intended visual structure as shown in last three columns of Table \ref{tab:wrong}. Such inconsistencies underscore the necessity of enhancing the model's capacity to generalize across diverse aspects of visual composition. 

\section{Conclusion and Future Work}
\label{sec:conclusion}
In this work, we address a critical challenge in text-to-image synthesis: the accurate interpretation and visual representation of complex prompts involving multiple objects and intricate spatial relationships. 
Our proposed approach, grounded in scene graph structures, constitutes a substantial advancement over prior methods through the integration of the ASQL Conditioner, which generates visual conditioning signals using a lightweight language model and directs the diffusion-based generation process via inference-time optimization. Through wide-ranging experiments, we demonstrate the capacity of our model.
Consequently, future research directions should focus on augmenting the model's architecture to improve its ability to accurately interpret and render complex compositional instructions, thereby bridging the gap between textual descriptions and their visual realizations. 


\bigskip

\bibliography{aaai2026}

\end{document}